\documentclass{article}

\usepackage{arxiv}

\usepackage[utf8]{inputenc} 
\usepackage[T1]{fontenc}    
\usepackage{hyperref}       
\usepackage{url}            
\usepackage{booktabs}       
\usepackage{amsfonts}       
\usepackage{nicefrac}       
\usepackage{microtype}      
\usepackage{lipsum}
\usepackage{authblk}

\usepackage{wrapfig}
\usepackage{balance}
\usepackage[numbers]{natbib}
\usepackage{float} 
\usepackage{times}
\usepackage{epsfig}
\usepackage{graphicx}
\usepackage{amsmath, amsfonts, amssymb}
\usepackage{bm}
\usepackage{algorithm}
\usepackage{algorithmicx}
\usepackage{comment}
\usepackage{enumitem}
\usepackage{fullpage}
\usepackage{tikz}
\usepackage{pifont}
\usepackage{amsthm}
\usepackage{subcaption}
\usepackage[table]{xcolor}

\usepackage{mathtools}
\usepackage{multirow}
\usepackage[noend]{algpseudocode}
\usepackage{booktabs}
\usepackage[export]{adjustbox}
\usepackage{url}

\usepackage{xcolor}
\hypersetup{
    colorlinks=true,
    linkcolor=blue,
    citecolor=blue,
    urlcolor=blue
}

\usepackage[textsize=scriptsize,backgroundcolor=green]{todonotes}

\usepackage{xspace}
\newcommand*{\eg}{\textit{e.g.},\@\xspace}
\newcommand*{\ie}{\textit{i.e.},\@\xspace}
\newcommand*{\vs}{\textit{vs.}\@\xspace}
\newcommand*{\wrt}{\textit{w.r.t.}\@\xspace}
\makeatletter
\newcommand*{\etc}{%
	\@ifnextchar{.}%
	{\textit{etc}}%
	{\textit{etc.}\@\xspace}%
}
\makeatother
\algtext*{EndWhile}
\algtext*{EndFor}
\algtext*{EndIf}
\algdef{SE}[DOWHILE]{Do}{doWhile}{\algorithmicdo}[1]{\algorithmicwhile\ #1}%

\makeatletter
\def\BState{\State\hskip-\ALG@thistlm}
\makeatother

\title{A Scalable MTL Framework for Modeling Demographic Disagreements in Annotation Tasks}

\author[1]{\textbf{Soumyajit Gupta}}
\author[2]{\textbf{Maria De-Arteaga}}
\author[3]{\textbf{Matthew Lease}}
\affil[1]{Computer Science, School of Computing, University of Texas at Austin}
\affil[2]{Universitat Ramon Llull, ESADE}
\affil[3]{Information, School of Computing, University of Texas at Austin}
\affil[ ]{\texttt{\small 
\{$^1$smjtgupta,$^3$ml\}@utexas.edu $^2$dearteaga@mccombs.utexas.edu}}

\begin{document}

\maketitle

\begin{abstract}

Annotators from different demographic backgrounds often disagree in systematic ways due to varying lived experiences, cultural contexts, and their relationship to the content. When such disagreement occurs, one group's perspective may sometimes warrant prioritization because they hold an epistemically advantaged standpoint or bear greater stake in the outcome. For example, when assessing toxicity of posts on social media, it may be important to prioritize a targeted group's viewpoint. We propose a \textit{Stakeholder-Aware Joint Multi-Task Learning} (\textbf{SAJ-MTL}) framework to jointly model demographic disagreement both across- and within- annotator groups. The model uses conditional gradient routing to handle sparsity induced by group-conditional labeling, and parameterization supports customized weighting of groups in different task settings. SAJ-MTL improves group-level performance in toxicity detection through better context-aware understanding of demographic-sensitive content. Only annotator group-membership information is required, which is more practical than existing methods requiring more privacy-invasive individual-level annotator details. Our scalable architecture is also computationally efficient to support large-scale deployment.

\end{abstract}

\section{Introduction}

Across many machine learning tasks, annotator judgments vary 
by demographic group \citep{uma2021learning}.
Such disagreements depend not only on annotator demographics, but also on the critical relationship between the annotator identity and the group for whom the content is most relevant or familiar~\citep{fazelpour2022diversity}. In medical diagnosis, Black and Latinx physicians in the US are historically more likely to serve in predominantly Black and Latinx communities~\citep{komaromy1996role}, and therefore may develop more expertise on the conditions that prevail in these communities and on how symptoms present. 
In a different domain of annotating online discourse, the sentiment of the phrase ``quite good'' depends on whether the speaker is British or American, as well as the annotator's familiarity with the idioms~\citep{bing2012sentiment}. From stove design \citep{gaviria2022dollar} to content moderation \citep{vaccaro2021contestability}, judgment decisions are often highly context-dependent and significantly influenced 
by an annotator's relationship to the content. Thus, when annotating for such highly subjective tasks, one must not only be cognizant and intentional about the relation between annotators and content, but when faced with disagreement, prioritizing certain viewpoints often becomes particularly essential in practice. 


%
Early crowdsourcing work predominantly treated disagreement as annotator noise to be minimized, collapsing multiple judgments into a single ``ground truth'' via majority vote~\citep{snow2008cheap,klebanov2009squibs,nowak2010reliable}. However, there is growing recognition that subjective annotation tasks \citep{tian2012learning,Nguyen16-hcomp} such as toxicity detection \citep{sap2019risk}, stance identification \citep{hardalov2022survey}, and sentiment analysis \citep{carvalho2023importance} are shaped by genuine differences in how demographic groups interpret content. Thus, a growing literature now argues that disagreement encodes meaningful social variation, and choices for how to handle disagreement are value-laden \citep{Frenda2024-qt}. \citet{fazelpour2022diversity} adopt a standpoint theory lens~\citep{intemann201025} to posit that the relationship between annotator identity and the content being annotated is a crucial dimension through which sociocultural identity and diversity is relevant to the machine learning pipeline. Empirical evidence lends support to this perspective. For example, \citet{davani2022dealing} empirically demonstrate that majority aggregation systematically erases minority viewpoints, a concern also raised on conceptual grounds by \citet{noble2012minority}, while \citep{gordon2022jury,sap2022annotators,orlikowski2023ecological} show that disagreement often reflects important structured group- and individual-level differences. 

While some works have tied annotator identity directly into models, this is typically done broadly as an auxiliary signal, not as a principled source of epistemic variation~ \citep{zhan2019learning,rottger2022two,kanclerz2022if,van2024annotator}. 
Most existing methods considering annotator identity do not fully account for how the annotator-content relationship influences subjective judgments. 
This gap has been acknowledged in the literature; \citet{binns2020apparent} highlights that although some perspectives carry greater epistemic authority in fairness-critical settings, current modeling pipelines lack capabilities to explicitly operationalize such prioritization.  \citet{fleisig2023majority}, a notable exception, attempts richer modeling of annotator variation. To do this, however, it requires fine-grained individual-level data or surveys, which raises significant privacy concerns and is often infeasible because such fine-grained data is rarely available. 
\textbf{Overall, we perceive an unmet need for a framework that:} \textbf{(1)} models both across-group and within-group disagreements explicitly; \textbf{(2)} enables dynamic, context-dependent viewpoint prioritization; and \textbf{(3)} remains computationally tractable without requiring privacy-invasive individual annotator-level features. This modeling space remains underexplored, 
lacking architectures that embed disagreement and socially grounded prioritization into the predictive pipeline itself.

To address these challenges, we propose a novel multi-task learning (MTL) architecture, Stakeholder-Aware Joint MTL (\textbf{SAJ-MTL}), that explicitly models interactions between annotator and target demographics to model context-dependent weights on socially-grounded perspectives. SAJ-MTL treats annotator–target relationships as primary signals rather than incidental metadata. In regard to 
challenge \textbf{(1)} above, instead of smoothing away disagreement, the model captures both across-group and within-group disagreements via two protocols: Across-Groups (AG) \vs Across-and-Within-Groups (AWG). By conditioning model predictions on annotator and target identities, the framework preserves multiple perspectives rather than collapsing them into a single label. As for challenge \textbf{(2)}, SAJ-MTL enables dynamic viewpoint prioritization by allowing the model to integrate weighting functions that reflect epistemic authority, democratic considerations of representation, or task-specific priorities, rather than assuming a single canonical perspective. Finally, for challenge \textbf{(3)}, SAJ-MTL 
achieves tractable computation by factorizing annotator and target representations into an additive MTL architecture. Avoiding the quadratic explosion of per-pair modeling enables the framework to handle many stakeholder groups without sacrificing predictive performance and memory stagnation.



We evaluate SAJ-MTL on 
online toxicity detection, 
a consequential setting wherein demographic disagreements directly shape annotation outcomes. Toxicity judgments are deeply subjective, tied to lived experience, historical context, and group-specific exposure to harm. As prior work shows, language that appears innocuous to one community may be offensive when directed toward another \citep{sap2019risk,davidson2019racial}. Moreover, when annotators hold negative beliefs about a group, they are less likely to label speech targeting that group as hateful~\citep{davani2023hate}. Toxicity datasets 
frequently employ majority-vote aggregation, which suppresses minority-group opinions and systematically biases models away from those most affected by harmful content \citep{davani2022dealing}. By considering the relationships between annotator and target demographics, SAJ-MTL produces a more faithful representation of how different communities perceive harm and enables viewpoint prioritization that centers the perspective of those at risk of being harmed. Evaluation 
on two toxicity datasets show SAJ-MTL improves context-aware understanding of demographically sensitive content and improves prediction. 


\textbf{Contributions:} 
\textbf{1)~Joint Stakeholder Modeling}. SAJ-MTL 
explicitly models demographic interactions between annotator, target and content, with generality allowing broad use across diverse subjective annotation tasks with multiple stakeholders. The model dynamically assesses and weighs different, socially-grounded perspectives given specific context rather than an averaged viewpoint. Conditional gradient routing supports sparse samples per task. 
\textbf{2)~Disagreement-Aware Protocols}. We describe two protocols to systematically model annotator disagreements and preserve diverse viewpoints erased by majority-vote aggregation: Across-Groups (AG) \vs Across-and-Within-Groups (AWG). 
\textbf{3)~Scalable MTL}. SAJ-MTL reduces scaling complexity from multiplicative to additive \wrt group cardinality of annotator and target, ensuring feasibility and cost-effectiveness for 
applications with many demographic groups.
\textbf{4)~Better Toxicity Detection}. When applied to toxicity detection, SAJ-MTL models how different annotator–target groups perceive harmful language, yielding predictions that respect demographic variation, leading to more context-sensitive toxicity judgments and reduced fairness disparities across groups.
\textbf{5)~Reproducibility}. We will share our source code via Github,
including architecture code and detailed experimental protocols. 
\section{Related Work}

We connect research in machine learning (ML) and natural language processing (NLP) on annotator disagreement, content relationships, and scalable MTL.



\textbf{Tasks with Annotator Disagreements.} 
Crowdsourcing and annotation work has repeatedly found structured heterogeneity in labels that correlates with annotator attributes \citep{uma2021learning}. 
Early work explicitly modeled annotator reliability and systematic biases \citep{dawid1979maximum,whitehill2009whose} while later work extended this to jointly learn annotator confusion matrices with classifiers \citep{raykar2010learning}. 
Even tasks with seemingly objective ground truth are labeled differently depending on the background, expertise, familiarity and lived experiences of the annotators, which 
is well-documented across domains. 
Judgments about stove design are highly subjective, 
with ideal forms varying significantly across the world and socioeconomic groups \citep{gaviria2022dollar}. In medical diagnosis, decisions can vary based on a clinician's training, experience, and normative practices of their professional environment \citep{ball2015improving}. Radiologists trained in different clinical settings may legitimately disagree on whether a faint shadow in an X‑ray signals pathology or artifact \citep{irvin2019chexpert}. In cardiology, recognized symptoms of a heart attack, such as crushing chest pain radiating down the left arm, are historically based on presentations in White male patients, but 
women and Black patients often present with different or subtler symptoms 
\citep{hilliard2020myocardial}. 
These examples 
motivate models that preserve annotator-level variation 
\citep{davani2022dealing}. 



\textbf{Modeling Annotator-Content Relationship.} Recent works model different dimensions of the annotator–content relationship directly in the modeling pipeline \cite{uma2021learning}. Jury learning \citep{gordon2022jury} treat annotators as stakeholders and learns to predict annotator-specific labels.
Personalized classifiers \cite{zhan2019learning} adapt predictions to an individual annotators' sensitivity using per-user annotation history, while noise-aware multi-annotator frameworks \cite{kanclerz2022if} estimate per-annotator reliability to denoise labels. These approaches capture individual preference, but none of these models the relationship between demographics of annotators and of another stakeholder (e.g. target) as context-dependent variation.
Most directly related to our work, \citet{fleisig2023majority} model annotator disagreement for subjective tasks by conditioning on annotator and target identity, serving as a precursor to our approach and a primary baseline.
While prior work demonstrates the value of conditioning on annotator identity, in practice it can be data-hungry or privacy-invasive by requiring per-annotator surveys or identifiers. 
%
However, \citet{sachdeva2022measuring} show that annotator behavior in subjective tasks like toxicity prediction can be efficiently modeled by decomposing judgments into independent latent traits (see 
\textbf{Appendix \ref{app:irt-linkage}} for a psychometric interpretation of SAJ-MTL as a neural generalization of identity-sensitive IRT). Our work contributes to this body of literature by proposing a scalable multi-task learning approach that models the relationship between annotator and target demographics.


\textbf{Demographic Bias and Fairness in NLP.} 
Models trained on aggregated annotations are well-known to reproduce and even amplify social biases. \citet{sap2019risk,davidson2019racial} show that marginalized perspectives can be systematically underrepresented when datasets collapse annotations. 
\citet{dixon2018measuring} highlight disparities in toxicity detection across demographic groups. Studies underscore the importance of fairness-aware modeling approaches that explicitly consider group-level variation.




\textbf{Scalability in MTL.} Scaling MTL models to many tasks (annotator $\times$ target heads in our work) is 
challenging. Naive per-task heads typically leads to multiplicative parameter growth in both memory and runtime \citep{caruana1993multitask}, so recent architectures emphasize parameter-efficiency and sparse conditioning. Adapter modules \citep{houlsby2019parameter} and adapter fusion \citep{pfeiffer2021adapterfusion} allow compact per-task updates, while Low-rank adaptation \citep{hu2022lora} and prompt/prefix tuning \citep{lester2021power} keep the per-task footprint small. Mixture-of-Experts \citep{shazeer2017outrageously,fedus2022switch} uses learned sparse routing to provide additional model capacity with controlled FLOPs, while multi-gate MoE \citep{ma2018modeling} and PLE \citep{tang2020progressive} propose task-specific gating for MTL. 
Techniques for gradient conflict mitigation \citep{yu2020gradient,chen2018gradnorm} address optimization interference among tasks and remain an active area of research. 

\section{Methods}

We now formalize the problem of demographic-aware annotation and introduce our Stakeholder-Aware Joint Multi-Task Learning (SAJ-MTL) framework. After highlighting inherent scalability issues with a naive multi-task formulation, we detail our scalable factorized architecture designed to efficiently capture both annotator and target group conditionality, labeling protocols, training objective and viewpoint prioritization scheme.

\subsection{Notation \& Problem Formulation}

We formalize the annotation setting as follows. Let $X$ denote the space of input samples (\eg text comments), which we map to a structured feature representation via a text encoder (\eg BERT~\citep{devlin2018bert}, RoBERTa~\citep{liu2019roberta}, \etc). Let $Y$ denote the label space (\eg toxicity score). Each comment is associated with a structured feature vector $x \in X$, an annotation label $y \in Y$, and metadata about the annotator-group and the target-group (mentioned or implicated in the content). We denote $A=\{a_1,a_2,\ldots,a_{|A|}\}$ as the set of annotator demographic groups, and $T=\{t_1,t_2,\ldots,t_{|T|}\}$ as the set of target demographic groups, with cardinality as $|A|$ and $|T|$ respectively.

Each training instance $i \in N$ is a tuple $(x_i,a_i,t_i,y_i)$, where $x_i \in X$ is the  content (feature vector of post), $a_i \in A$ is the annotator's demographic group, $t_i \in T$ is the target's demographic group, and $y_i \in Y^{\{0,1\}}$ is the binary judgment noting the label given by annotator $a_i$ to instance $x_i$ when considered in relation to target $t_i$ (\eg is  a post hateful towards a particular demographic). We assume the observed label is drawn from a conditional distribution $y_i \sim P(y \mid x_i,a_i,t_i)$, where both annotator- and target-group identities influence the group-conditioned interpretation of the content. This explicitly departs from both traditional single-truth formulations that assume $y_i \sim P(y \mid x_i)$ alone governs labeling, and stakeholder-aware formulations that condition predictions on the content and a single stakeholder at a time.

\subsection{Naive Extension of Conditional MTL} \label{sec:CondMTL}

\noindent
\begin{wrapfigure}{r}{0.49\textwidth} 
    \centering
    \vspace{-3em}
    \includegraphics[width=\linewidth]{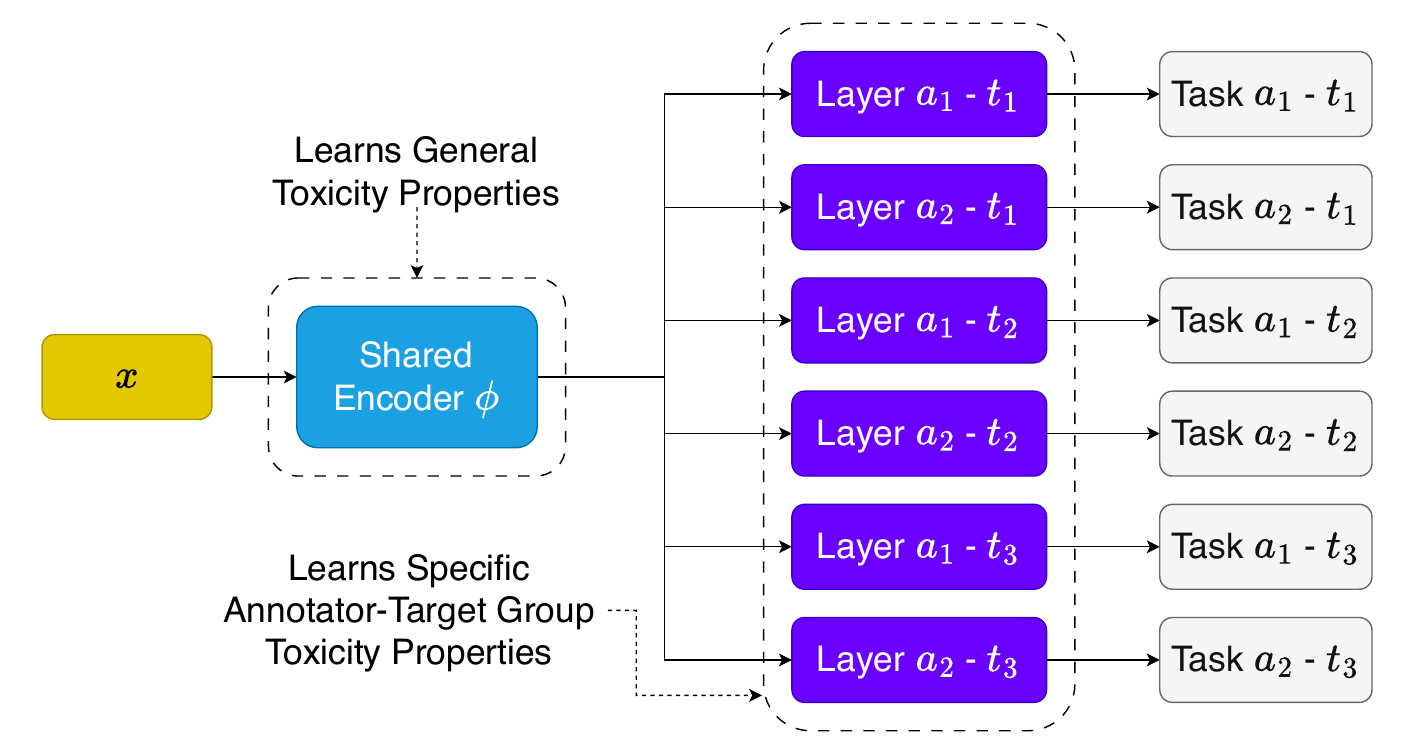}
    \vspace{-2em}
    \caption{Naive All-Pairs SAJ-MTL. Each annotator-target pair $(a,t)$ is modeled as a separate task 
    and dedicated head,  allowing direct learning of pairwise interactions but yielding multiplicative parameter growth and 
    data sparsity for underrepresented pairs.}
    \vspace{-1em}
    \label{fig:naive}
\end{wrapfigure}
\citet{gupta2023same}'s CondMTL combines a shared encoder with multiple task-specific prediction heads. One simple way to model demographic variation would be to extend CondMTL with conditional gradient routing, as shown in \textbf{Fig.~\ref{fig:naive}}, with each prediction head corresponding to 
a combination of an annotator group $a \in A$ and a target group $t \in T$. 
While this structure allows the model to learn localized annotator–target interactions directly, it suffers from severe scalability and data sparsity issues. In this naive setup, the number of task-specific parameters grows multiplicatively at a rate of $\mathcal{O}(|A| \times |T|)$. Many demographic pairings in real-world settings, including MHS \citep{sachdeva2022assessing} and KET \citep{kumar2021designing}, are quite sparse (\eg $\sim$ 1\% of data). This makes it nearly impossible for dedicated all-pair heads to converge effectively without overfitting. 

\begin{figure}[ht]
   \centering
   \vspace{-0.5em}
\includegraphics[width=0.8\linewidth]{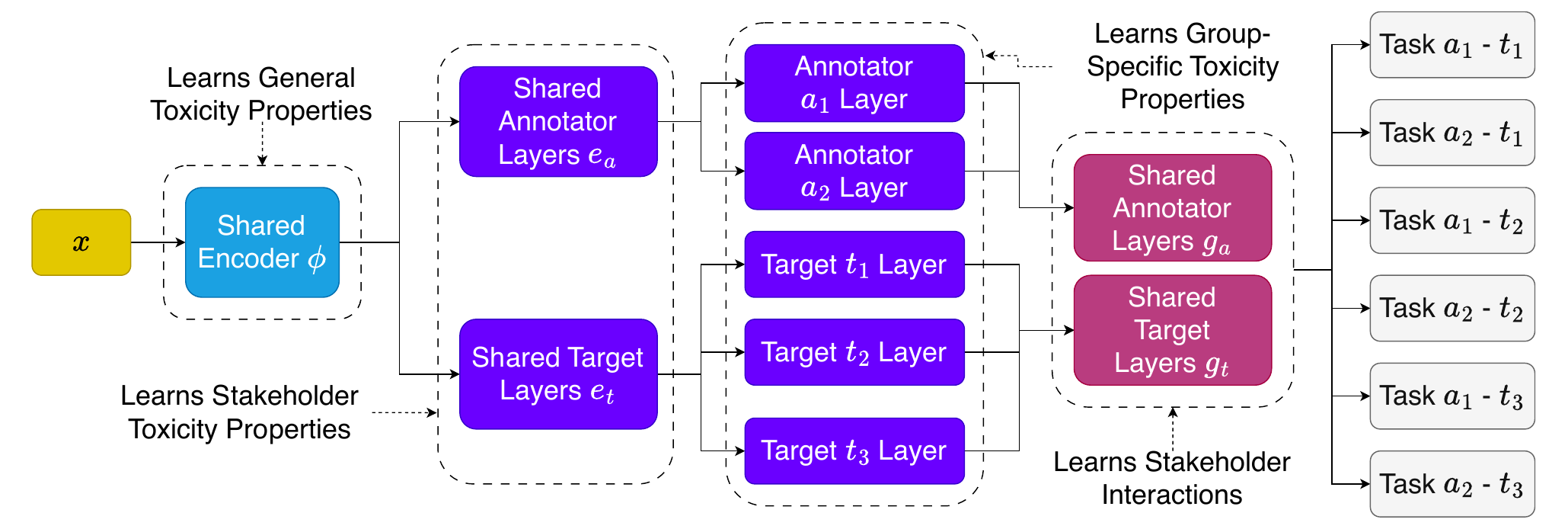}
    \vspace{-1em}
    \caption{Scalable SAJ-MTL. Annotator and target effects are factorized into independent modules and combined additively. This enables efficient modeling of annotator–target interactions while reducing complexity to linear scaling, improving both parameter efficiency and generalization in sparse settings.}
    \vspace{-1.25em}
    \label{fig:scalable}
\end{figure}

To overcome combinatorial explosion and data sparsity issues, our improved scheme must decouple the annotator-specific variations from the target-specific contexts, while still allowing the model to account for the relationship between them. By factorizing these components, as shown in \textbf{Fig.~\ref{fig:scalable}}, we can achieve a scalable architecture that learns these effects independently and composes them additively. This ensures the model scales efficiently at $\mathcal{O}(|A| + |T|)$, sharing general representation knowledge while still effectively modeling the complex, subjective interactions between the annotator and 
the targeted group. 


\subsection{Modeling Annotator-Target Conditionality} \label{sec:arch}

We model the subjective prediction of input $x$, annotated by demographic group $a$ and referring to target group $t$, using separate representations for the input and each stakeholder identity. Specifically, the shared encoder (with parameters $\phi$) produces a content embedding $h_{\phi}(x) \in \mathbb{R}^d$, while learnable embedding layers produce annotator embeddings $e_{a} \in \mathbb{R}^{d_a}$, and target embeddings $e_{t} \in \mathbb{R}^{d_t}$, corresponding to the demographic groups $a$ and $t$, respectively. These three embedings are then combined through the proposed conditional architecture yielding a stakeholder-conditioned representation:
\begin{equation}
    z(x,a,t) = f_{cond}(h_{\phi}(x),e_a,e_t) \label{eq:fcond}
\end{equation}
where $f_{cond}$ is a conditional parameterization function that composes the three components into a unified latent representation. The function may take various forms depending on modeling priorities. The fused representation is then passed through branch-specific module heads $g_a,g_t$, yielding conditional predictions:
\begin{equation}
    f_{\theta}(y \mid x,a,t) = \text{softmax} \, [ \, g_a(z(x,a,t)) + g_t(z(x,a,t)) \,]
\end{equation}
This enables annotator-target specific losses to backpropagate through the shared encoder $h_{\phi}$ and the relevant group-specific embeddings $e_a$, $e_t$. The conditional structure ensures branch-specific updates: for a given example $(x,a,t,y)$, gradients only adjust the corresponding annotator and target embeddings, along with the shared parameters. This selective mechanism allows the model to specialize efficiently for specific annotator-target contexts without interference from unrelated pairs, and is less prone to data sparsity issues.

\begin{figure}[ht]
    \centering
    \includegraphics[width=0.9\linewidth]{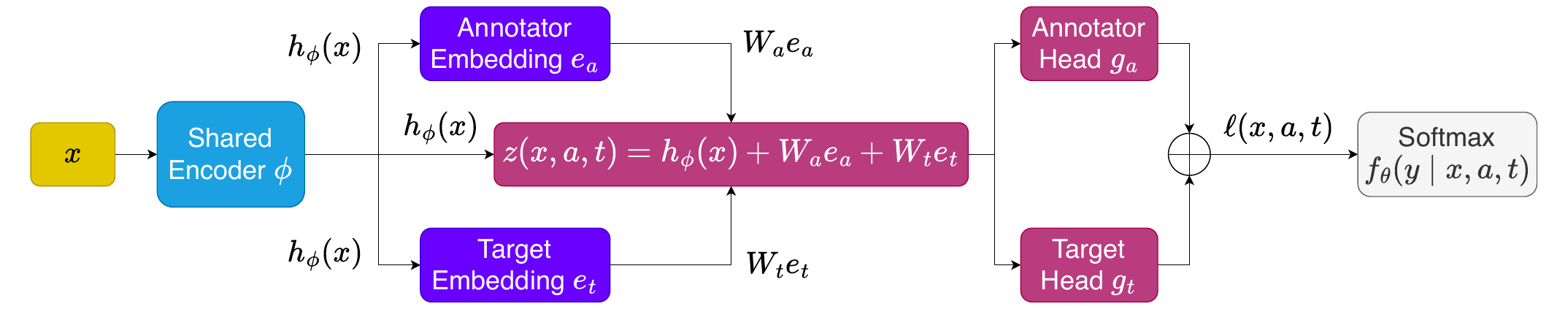}
    \vspace{-0.5em}
    \caption{\textbf{Overview of the SAJ-MTL architecture}. Input vector $x$ is encoded using a shared encoder $h_{\phi}(x)$. Annotator-group and target-group embeddings $(e_a,e_t)$ are projected via lightweight linear modules $(W_a,W_t)$ and fused additively with the shared representation to form the conditional representation $z(x,a,t)$. Two parallel prediction heads, annotator-specific head $g_a$ and target-specific head $g_a$, map this representation to logits that are summed to produce the annotator–target–conditioned prediction $f_{\theta} (y \mid x,a,t)$. During training, gradients update only the shared encoder and the annotator/target modules corresponding to the active pair $(a,t)$. At inference, predictions may be aggregated using viewpoint prioritization weights $\pi(a,t)$ enabling socially grounded or policy-driven weighting of perspectives (see Section~\ref{sec:viewpoint}).}
    \label{fig:sajmtl}
    \vspace{-1em}
\end{figure}

\subsection{Architectural Overview}

SAJ-MTL implements annotator–target conditionality 
with each annotator–target pair corresponding to a logical task. Parameters are composed from independent annotator- and target-specific modules, rather than instantiated explicitly (see \textbf{Fig.~\ref{fig:sajmtl}}). Whereas conventional MTL tasks are predefined and static, SAJ-MTL tasks are conditionally activated based on annotator and target metadata for each labeled instance. 
We implement the conditional parameterization function $f_{cond}$ as a linear additive function to empirically balance expressiveness and computational efficiency.
\begin{equation}
    z(x,a,t) = h_{\phi}(x) + W_a e_a + W_t e_t \label{eq:fcond-add}
\end{equation}
Given the conditional representation $z(x,a,t)$, SAJ-MTL assigns a distinct prediction head to each annotator group $a$ and a distinct prediction head to each target group $t$. Formally, let $g_a, g_t: \mathbb{R}^d \rightarrow \mathbb{R}^k$ denote the annotator-specific and target-specific heads, respectively. The final logits for an instance labeled by annotator $a$ about target group $t$ are obtained by combining both heads with the conditional representation:
\begin{equation}
    \ell(x,a,t) = g_a(z(x,a,t)) + g_t(z(x,a,t))
\end{equation}
The additive structure allows annotator-related semantic shifts and target-related contextual shifts to contribute independently and interpretably, unlike 
learning a single classifier conditioned on metadata tokens \citep{fleisig2023majority}, which entangles annotator and target effects for less explicit disagreement modeling.

\subsection{Training Objective and Optimization}

Training on fine-grained annotations rather than aggregated labels 
preserves and learns from structured disagreement. For each labeled instance $(x_i,a_i,t_i,y_i)$, annotator-target–conditioned logits are produced where the conditional representation $z(x_i,a_i,t_i)$ includes 
shared content features and group-specific effects.
\begin{equation}
    \ell(x_i,a_i,t_i) = g_{a_i}(z(x_i,a_i,t_i)) + g_{t_i}(z(x_i,a_i,t_i)) \label{eq:ann-tar-logits}
\end{equation}
The model is trained using the conditional negative log-likelihood:
\begin{equation}
    \mathcal{L}(\theta) = -\frac{1}{N} \sum_{i=1}^N \log f_{\theta} (y_i \mid x_i,a_i,t_i) = -\frac{1}{N} \sum_{i=1}^N \log \text{softmax} \, (\ell(x_i,a_i,t_i))_{y_i} \label{eq:pi}
\end{equation}
A key feature of SAJ-MTL is selective gradient routing. For each $(x_i,a_i,t_i,y_i)$, gradient updates are applied only to: i) the shared content encoder parameters $\phi$; ii) the annotator embedding $e_{a_i}$ and its head $g_{a_i}$; iii) the target embedding $e_{t_i}$ and its head $g_{t_i}$; while all other annotator and target modules remain unchanged. This yields branch-specific backpropagation and ensures that each demographic group shapes its own subspace of the representation, allowing the model to learn distinct patterns without interference from unrelated groups.

As a conditional gradient routing model, SAJ-MTL encounters the typical challenge of gradient interference across tasks. However, because SAJ-MTL decomposes annotator and target effects additively, the parameter space naturally reduces conflicting gradients. One may further incorporate MTL optimization techniques such as gradient normalization (GradNorm \citep{chen2018gradnorm}), gradient projection (PCGrad \citep{yu2020gradient}) or multi-objective balancing (MGDA \citep{sener2018multi}), to modulate interactions between annotator and target updates. In practice, we find that the additive factorization is stable under standard optimizers such as AdamMax \citep{kingma2014adam} without requiring explicit multi-task gradient control. To prevent overfitting on small groups and ensure meaningful separation between annotator and target effects, SAJ-MTL applies standard $L_2$ regularization on head parameters, dropout within the dense modules, and norm constraints on the embeddings $e_a,e_t$ to prevent representational collapse.
\vspace{-0.5em}

\subsection{Viewpoint Prioritization Mechanism} \label{sec:viewpoint}

Subjective annotation tasks often require privileging certain perspectives over others, especially in contexts where lived experience or group-specific harm confers epistemic authority. To support such socially grounded decision-making, SAJ-MTL incorporates a viewpoint prioritization mechanism that modulates predictions based on the relationship between annotator and target groups. This allows the model to flexibly reflect platform policies, application goals, or normative choices about whose perspective should guide evaluation. 

Given the conditional prediction function $f_\theta(y \mid x, a, t)$, 
a prioritization weight $\pi(a,t)$ 
specifies the 
influence of annotator group $a$ 
when assessing content involving target group $t$. The prioritization function: $\pi: A \times T \rightarrow [0,1]$ is constrained to satisfy $\sum_{a \in A} \pi(a, t)=1$ for each target group $t$. At inference time, the model produces an aggregate prediction by marginalizing over annotator perspectives under the chosen weighting scheme:
\begin{equation}
    \hat{y} = P(y \mid x,t) = \arg\max_y \sum_{a \in A} \pi(a,t) f_\theta(y \mid x,a,t)
\end{equation}
This general formulation allows SAJ-MTL 
to implement a variety of prioritization strategies  
proposed in prior literature on stakeholder-aware and participatory modeling \citep{kruglanski2005says,sorensen2024position,kallus2018balanced,yeh1999task}. Below we relate $\pi(a,t)$ to common prioritization strategies. Refer to \textbf{Section~\ref{sec:metrics}} for the choice and reasoning relevant to our own empirical experiments.
\begin{itemize}[leftmargin=*, itemsep=1pt]
    \item \textit{Epistemic authority weighting}: $\pi(a,t)$ increases for examples where $a=t$, reflecting that members of a targeted group often have the strongest basis for judging harm.
    \item \textit{Pluralistic weighting}: $\pi(a,t)$ is uniform across groups, where diverse perspectives are modelled equally.
    \item \textit{Policy-driven weighting}: $\pi(a,t)$ reflects community guidelines, safety thresholds, or regulatory constraints.
    \item \textit{Task-specific weighting}: domain expertise or annotator proficiency determines $\pi(a,t)$.
\end{itemize}

%

Note that viewpoint prioritization \textbf{only affects aggregation} and not the architecture or training procedure. The conditional model $f_\theta$ continues to learn fine-grained differences in how annotators interpret content, while $\pi(a,t)$ determines which perspectives are emphasized once the model is applied. This decoupling 
allows SAJ-MTL to be used in diverse settings without embedding implicit choices into the learned parameters. This also supports fairness-aware evaluation: by varying $\pi(a,t)$, one can explore how model performance or disparities shift under other prioritization schemes, highlighting whose perspectives current systems advantage or ignore. This is especially valuable in domains like toxicity detection, where harm is not uniformly experienced and where platform governance may require accounting for group-specific viewpoints.



\section{Experimental Setup}

Our experiments consider two scenarios: a controlled setting in which target groups, $t$, are known (human-annotated) \vs a more realistic setting in which targets must be noisily inferred. We first introduce these scenarios and corresponding datasets, followed by evaluation metrics, baselines, and data pipeline protocols used in our experiments. Additional  
implementation details 
can be found in \textbf{Appendix \ref{app:details}}.

\subsection{Evaluation Scenarios and Datasets}

We evaluate SAJ-MTL on two datasets for toxicity and hate-related classification that include annotator demographics. Both datasets are well-suited for studying annotator disagreement, with multiple labels per instance and systematic variation across demographic groups. We distinguish two evaluation settings. First, the \textbf{Annotated-Target Scenario} (Section~\ref{sec:annotated-targets}) provides a controlled-setting with definitively known targets via human annotation, i.e. annotations indicating which demographic group is the target of a post. The Measuring Hate Speech (MHS) corpus \citep{sachdeva2022assessing} supports such evaluation. Practical deployment, however, requires target detection. This is the \textbf{Detected-Target Scenario}, a more realistic and noisy evaluation condition (Section~\ref{sec:detected-targets}). This particularly matters for SAJ-MTL, since the model conditions on target-group identity during inference. Together, these two settings enable us to assess both the upper-bound performance of the model under known target conditions \vs practical effectiveness in more realistic settings. For the target detection scenario, we detect targets using a custom detection module (see \textbf{Appendix \ref{app:noisy-oracle}}), then report results on MHS (ignoring human-annotated target labels) and on the KET dataset \citep{kumar2021designing} (which lacks human-annotated targets).


MHS contains posts sourced from YouTube, Twitter and Reddit. 
Although the dataset has seven demographic race groups, we keep four demographic groups (\textit{Asian}, \textit{Black}, \textit{Latinx} and \textit{White}), omitting three others (\textit{Middle-Eastern}, \textit{Native-American} and \textit{Pacific-Islander}) due to extremely sparse annotator–target pair instances. 
KET contains posts sourced from Twitter, Reddit, and 4chan, and we focus on four annotator groups based on race demographics: \textit{Asian}, \textit{Black}, \textit{Latino}, and  \textit{White}. Each annotator has marked the post as being non-toxic (0) or toxic (1-4), which we binarize as $y \in \{0,1\}$ for our formulation. Additional statistics on group frequencies and the distribution of targeted demographics are provided in \textbf{Appendix \ref{app:dataset}}.
\subsection{Baselines} \label{sec:baselines}

To evaluate the effectiveness of SAJ-MTL, we compare it against two representative baseline approaches that incorporate demographic information into toxicity classification models. These baselines reflect commonly used strategies for handling demographic metadata and multi-task structures in subjective annotation tasks.

\begin{wrapfigure}{r}{0.49\textwidth} 
    \centering
    \vspace{-1.5em}
    \includegraphics[width=0.9\linewidth]{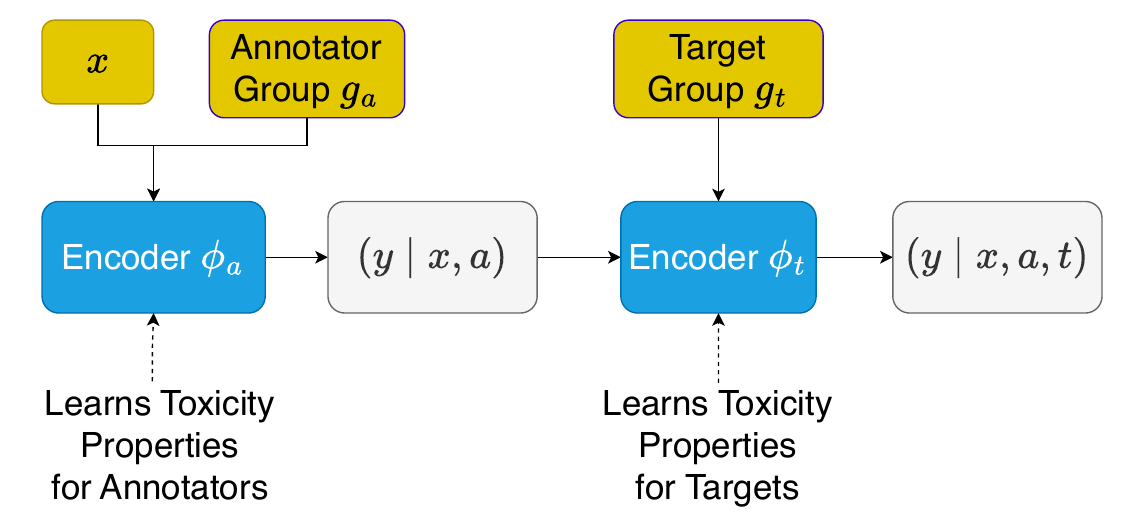}
    \vspace{-1em}
    \caption{TextAug Pipeline. The input text is internally predicted for toxicity based on the annotator group. An additional encoder is then used to predict conditioned on the target group as well.}
    \vspace{-1.0em}
    \label{fig:textaug}
\end{wrapfigure}
1. Text Augmentation (\textbf{TextAug}) \citep{fleisig2023majority}. 
The input text is augmented with tokens describing the annotator’s demographic group (see \textbf{Fig.~\ref{fig:textaug}}). 
The intuition is that the encoder can learn context-dependent representations conditioned on the annotator identity through attention mechanisms. The augmented input is processed by a shared encoder $h_{\phi}$, and the resulting representation is passed through a classification head to produce toxicity predictions. While this method allows demographic metadata to influence the learned encoded representation, it does not explicitly model structured relationships between annotator and target groups. The influence of demographic metadata remains entangled with the text representation, making it difficult for the model to learn distinct interaction patterns between stakeholders.

2. Conditional Multi-Task Learning (\textbf{CondMTL}) \citet{gupta2023same} (see \textbf{Section~\ref{sec:CondMTL}}). Each task corresponds to a target demographic group,  with the model learning a distinct classifier for toxicity directed at each group. It thus  captures only target-specific variation in toxicity and not the relational structure of who is judging what. This baseline provides a useful reference for assessing whether explicit modeling of annotator–target relations provides useful predictive signal beyond target-aware classification alone.



\subsection{Evaluation Metrics} \label{sec:metrics}


\textbf{Classification}: Given the inherent class imbalance and multi-group nature of the task, we primarily focus on macro-level metrics (averaging performance across groups) to ensure that performance on minority demographic groups is weighted equally to that of majority groups. For completeness, we report the micro-level (averaging performance across all predictions) metrics as well. Metrics include Precision, Recall and F1.

\textbf{Fairness}: To assess how well the model respects demographic variation and reduces systemic bias, we utilize fairness-specific metrics that track disparities in error rates. Specifically, we evaluate Equal Opportunity (EO) Gap \citep{hardt2016equality} as the maximum difference in TPR (Recall) between any two demographic groups, as specified in Eq.~\ref{eq:EO}. In toxicity detection, lower TPR indicates that hateful posts targeting a certain group are not flagged as such.
\begin{equation}
\label{eq:EO}
    \mathrm{Max.\ EO\ Gap} = \max_{g \in \{A, T\}} TPR_g - \min_{g \in \{A, T\}} TPR_g
\end{equation}


\textbf{Viewpoint Aggregation}: We instantiate a group-aligned epistemic authority scheme that assigns larger weights when annotator and target demographics coincide (\ie where $a=t$). This is motivated by prior work \citep{gordon2022jury,fleisig2023majority} suggesting that targeted group members may have greater contextual familiarity and lived experience for evaluating potentially harmful content directed toward them.

\subsection{Data Pipeline Protocols}

Subjective annotation datasets often contain multiple labels per instance that capture variation both across groups and within the same group. To formally model this, we consider two labeling strategies distinguishing how labels are indexed and used by SAJ-MTL: AG (Across-Groups) \vs AWG (Across-and-Within-Groups). This distinction is critical for learning differences in subjective data. The AG strategy focuses on disparate belief \ie modeling systematic differences in labeling behavior between distinct groups. In contrast, the AWG strategy addresses the ecological fallacy of assuming group-level trends apply uniformly to all individuals by explicitly modeling variation both across groups and within each group. This enables labels to capture the fact that not all members within a group necessarily agree. Such an approach enables a granular representation that avoids over-generalization while still capturing inter-group differences.

\textbf{AG (Across-Groups):} Each sample receives a single, group-consolidated label per group representing the group's average/majority perspective. Different label instances for the same sample therefore correspond strictly to different groups, treating each group's judgment as a cohesive unit. This approach captures systematic differences between groups but does not account for individual disagreement within a group.

\textbf{AWG (Across-and-Within-Groups):} A single sample can have multiple labels from within the same demographic group, in addition to labels from different groups. This allows the model to see several differing opinions from individuals in the same group. This captures both inter-group and intra-group differences, reflecting the rich spectrum of individual disagreement that exists inside any demographic category.

AG captures systematic differences across demographic groups by learning from a single representative label per group, whereas AWG additionally preserves disagreement among individuals within the same demographic group. Because within-group disagreement provides additional information about the diversity of stakeholder perspectives, we hypothesize that AWG will learn richer stakeholder-conditioned representations and consequently achieve better predictive performance than AG. Our experiments in \textbf{Section \ref{sec:results}} evaluate whether preserving this additional disagreement translates into measurable improvements.

\section{Results} \label{sec:results}

%

We evaluate SAJ-MTL along three complementary axes: 1) predictive performance; 2) fairness across demographic groups; and 3) computational scalability. 
Per Section~\ref{sec:metrics}, all results are computed using the \textit{epistemic authority} weighting function $\pi(a,t)$ (Section~\ref{sec:viewpoint}); corresponding to the strategy in TextAug baseline \citep{fleisig2023majority}, \ie $\pi(a,t)=1$, when group identities match ($a=t$), else 0. We compare SAJ-MTL against CondMTL and TextAug baselines (Section~\ref{sec:baselines}). 
Our main results use DistilBERT as text encoder; see \textbf{Appendix \ref{app:roberta}} foradditional results using RoBERTa-base. 

\subsection{Predictive Performance on MHS with Annotated-Targets} \label{sec:annotated-targets}

\textbf{Table \ref{tab:joint-summary-mhs}} reports classification results on the MHS dataset with human-annotated targets, and reports both micro- and macro-averaged metrics. 
SAJ-MTL consistently outperforms baselines, showing the benefit of explicitly modeling annotator–target relationships. In regard to SAJ-MTL variants, AWG labeling performs better than AG labeling, reaching a (macro/micro)-F1 of (0.65, 0.64), highlighting the 
benefit of modeling both across-group disagreements and within-group variation, which are lost when annotations are aggregated or treated as homogeneous within each demographic group. 
In terms of baselines, CondMTL, which models only the target demographic as a task, performs substantially worse across all metrics. This suggests that modeling the annotator group is critical for some subjective tasks such as toxicity detection, where judgments depend strongly on the relationship between the annotator and the content they are annotating. The inclusion of annotator demographics serve as an additional dimension for the model to learn a better toxicity boundary. Baseline TextAug performs better than CondMTL for most groups, but remains significantly below SAJ-MTL. This suggests that while text augmentation allows annotator metadata to influence the encoder representation, it does not provide explicit structural modeling of annotator–target interactions. Consequently, TextAug appears to struggle to disentangle how different communities interpret the same content. 


\begin{table}[h]
    \centering
    \vspace{-0.75em}
    \caption{Prediction results on MHS for the Annotated-Target scenario. Mean performance and standard deviation are reported over 5 runs. The top results are bolded. SAJ-MTL with Joint-AWG labeling is best.}
    \resizebox{0.65\textwidth}{!}{
    \begin{tabular}{r|c|cc|cc} \toprule
        Metric & Level & Joint-AWG & Joint-AG & TextAug & CondMTL \\ \midrule
        \multirow{ 2}{*}{F1 ($\uparrow$)} & Micro & \textbf{0.64 $\pm$ 0.02} & 0.61 $\pm$ 0.01 & 0.52 $\pm$ 0.03 & 0.50 $\pm$ 0.02 \\
        & Macro & \textbf{0.65 $\pm$ 0.01} & 0.64 $\pm$ 0.02 & 0.55 $\pm$ 0.03 & 0.49 $\pm$ 0.01 \\ \midrule
        \multirow{ 2}{*}{Precision ($\uparrow$)} & Micro & \textbf{0.71 $\pm$ 0.01} & 0.67 $\pm$ 0.02 & 0.63 $\pm$ 0.03 & 0.60 $\pm$ 0.02 \\
        & Macro & \textbf{0.75 $\pm$ 0.02} & 0.73 $\pm$ 0.01 & 0.66 $\pm$ 0.02 & 0.59 $\pm$ 0.01 \\ \midrule
        \multirow{ 2}{*}{Recall ($\uparrow$)} & Micro & \textbf{0.59 $\pm$ 0.02} & 0.57 $\pm$ 0.01 & 0.45 $\pm$ 0.03 & 0.43 $\pm$ 0.02 \\
        & Macro & \textbf{0.59 $\pm$ 0.01} & 0.58 $\pm$ 0.02 & 0.47 $\pm$ 0.02 & 0.42 $\pm$ 0.02 \\ 
        \bottomrule
    \end{tabular}
    }
    \label{tab:joint-summary-mhs}
    \vspace{-0.5em}
\end{table}

\begin{figure}[ht]
    \centering
    \includegraphics[width=0.6\linewidth]{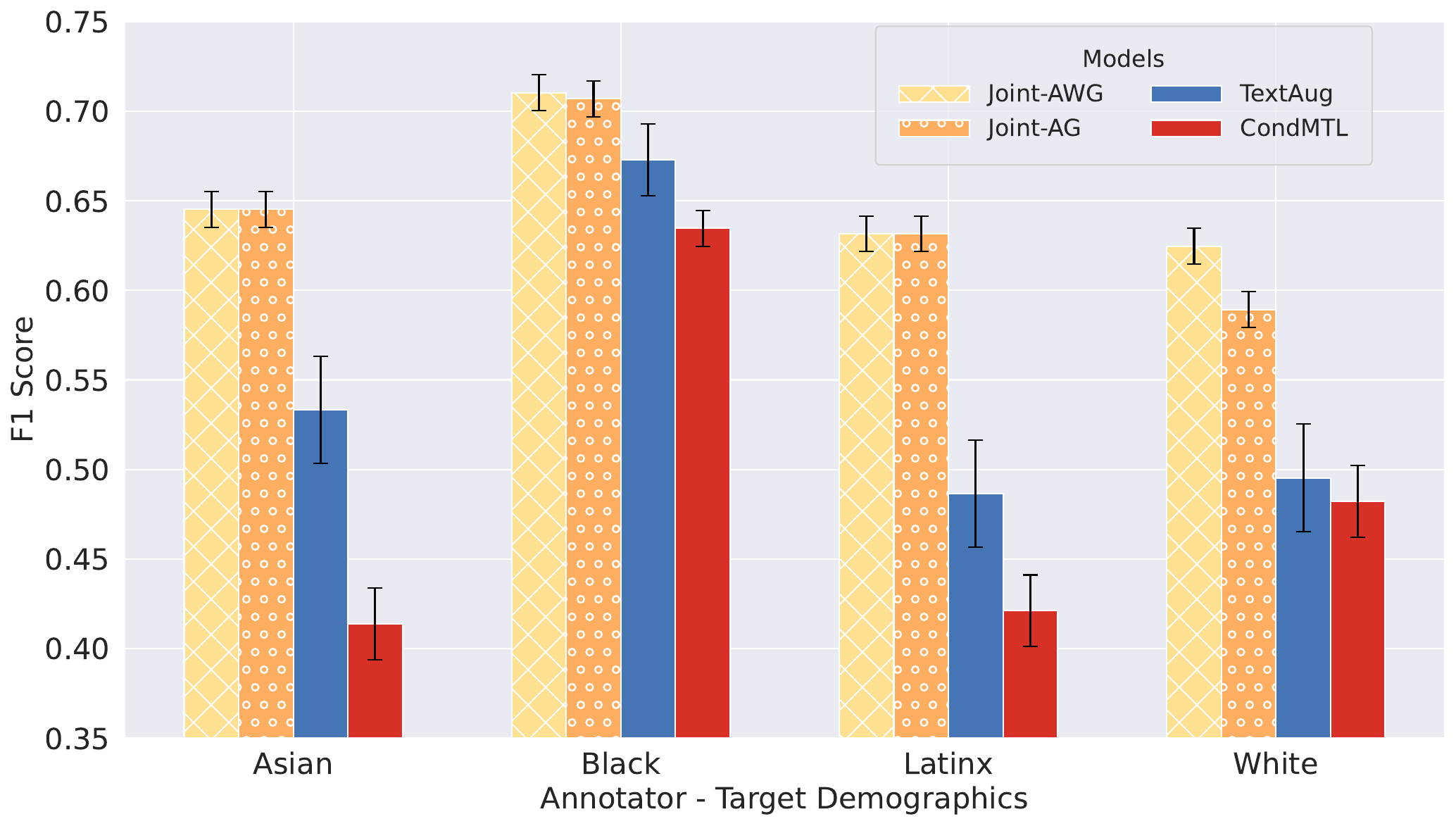}
    \vspace{-0.75em}
    \caption{F1 scores over Annotator - Target groups. Both of our Joint-AWG and Joint-AG models outperform the baseline models by better capturing the interaction between annotator and target demographics. 
    }
    \label{fig:f1-joint-mhs}
    \vspace{-0.75em}
\end{figure}

In sum, results affirm that annotator disagreement contains meaningful signal: modeling this can capture more nuanced interpretations of harmful language. \textbf{Fig.~\ref{fig:f1-joint-mhs}} analyzes performance across specific annotator–target demographic pairings under epistemic authority weighting, with AWG labeling showing the largest gains for groups with higher internal disagreement, especially for instances involving White annotators.



\begin{table}[h]
    \centering
    \vspace{-0.5em}
    \caption{Per-group recall (higher is better) for each demographic group, along with the corresponding Maximum Equal Opportunity (EO) Gap (\textbf{lower is better}). Max. EO Gap is the difference between the maximum and minimum recall across demographic groups. Both Joint-AWG and Joint-AG achieve the lowest EO Gap, indicating more equitable performance across demographic groups compared to other models.}
    \resizebox{0.7\textwidth}{!}{
    \begin{tabular}{l|cccc|c} \toprule
         \multirow{2}{*}{Model} & \multicolumn{4}{c|}{Recall} & Max. \\ 
         & Asian & Black & Latinx & White & EO Gap ($\downarrow$) \\ \midrule
         Joint-AWG & 0.53 $\pm$ 0.01 & 0.70 $\pm$ 0.01 & 0.55 $\pm$ 0.01 & 0.57 $\pm$ 0.01 & \textbf{0.17 $\pm$ 0.01} \\
        Joint-AG & 0.53 $\pm$ 0.01 & 0.70 $\pm$ 0.01 & 0.55 $\pm$ 0.01 & 0.54 $\pm$ 0.01 & \textbf{0.17 $\pm$ 0.01} \\
        TextAug & 0.42 $\pm$ 0.03 & 0.64 $\pm$ 0.02 & 0.41 $\pm$ 0.03 & 0.41 $\pm$ 0.03 & 0.23 $\pm$ 0.03 \\
        CondMTL & 0.32 $\pm$ 0.02 & 0.61 $\pm$ 0.01 & 0.36 $\pm$ 0.02 & 0.40 $\pm$ 0.02 & 0.29 $\pm$ 0.02 \\ 
        \bottomrule
    \end{tabular}
    }
    \vspace{-0.5em}
    \label{tab:eogap-mhs}
\end{table}

\textbf{Fairness}. We next examine whether modeling demographic disagreement into the framework improves fairness across groups, measured using the Equal Opportunity (EO) gap (\textbf{Section~\ref{sec:metrics}}) on MHS. 
\textbf{Table \ref{tab:eogap-mhs}} shows that both AG and AWG SAJ-MTL variants achieve the lowest EO gaps, with Joint-AWG and Joint-AG reducing the gap to 0.17 compared to 0.23 for TextAug and 0.29 for CondMTL. These improvements likely arise because SAJ-MTL explicitly models the relationship between annotator and target demographics. By learning separate annotator- and target-specific representations, the model can capture differences in how communities perceive harmful language without forcing a single universal decision boundary. Notably, the reduction in EO gap occurs without any fairness-specific optimization objective, suggesting that modeling disagreement itself can mitigate disparities that arise when minority perspectives are suppressed during label aggregation.

\subsection{Predictive Performance on MHS \& KET with Detected-Targets} \label{sec:detected-targets}

We present the predictive performance results of SAJ-MTL under a more realistic deployment setting in which target demographic identities are not assumed to be known a priori. Instead, target groups are predicted automatically using an auxiliary target detection module (see \textbf{Appendix \ref{app:noisy-oracle}}) and these predictions are used during both training and inference. This reflects practical moderation scenarios where demographic targets (e.g. whether a post is targeted towards Black people or LatinX people) must be inferred rather than provided as human annotations. We present results for MHS (ignoring the human-annotated target labels) and KET (which lacks target annotations) in \textbf{Table~\ref{tab:f1-scores-noisy-oracle}}. 
%



\begin{table}[h]
    \centering
    \vspace{-0.75em}
    \caption{Predictive performance on datasets for the Detected-Target scenario, averaged over five runs.}
    \resizebox{\textwidth}{!}{
    \begin{tabular}{r|c|cc|cc||cc|cc} \toprule
        Dataset &  & \multicolumn{4}{c||}{MHS} & \multicolumn{4}{c}{KET} \\
        Metric & Level & Joint-AWG & Joint-AG & TextAug & CondMTL & Joint-AWG & Joint-AG & TextAug & CondMTL \\ \midrule
        \multirow{ 2}{*}{F1 ($\uparrow$)} & Micro & \textbf{0.58 $\pm$ 0.02} & 0.56 $\pm$ 0.02 & 0.48 $\pm$ 0.04  & 0.45 $\pm$ 0.03 & \textbf{0.75 $\pm$ 0.03} & 0.72 $\pm$ 0.04 & 0.65 $\pm$ 0.02 & 0.62 $\pm$ 0.03 \\
        & Macro & \textbf{0.59 $\pm$ 0.01} & 0.57 $\pm$ 0.02  & 0.49 $\pm$ 0.03 & 0.45 $\pm$ 0.02 & \textbf{0.71 $\pm$ 0.02} & 0.68 $\pm$ 0.03 & 0.60 $\pm$ 0.04 & 0.56 $\pm$ 0.03 \\ \midrule
        \multirow{ 2}{*}{Precision ($\uparrow$)} & Micro & \textbf{0.68 $\pm$ 0.02} & 0.66 $\pm$ 0.01 & 0.58 $\pm$ 0.02 & 0.53 $\pm$ 0.01 & \textbf{0.77 $\pm$ 0.03} & 0.74 $\pm$ 0.03 & 0.68 $\pm$ 0.01 & 0.65 $\pm$ 0.04 \\
        & Macro & \textbf{0.73 $\pm$ 0.01} & 0.71 $\pm$ 0.01 & 0.58 $\pm$ 0.01 & 0.55 $\pm$ 0.01 & \textbf{0.71 $\pm$ 0.02} & 0.69 $\pm$ 0.03 & 0.63 $\pm$ 0.03 & 0.58 $\pm$ 0.05 \\ \midrule
        \multirow{ 2}{*}{Recall ($\uparrow$)} & Micro & \textbf{0.51 $\pm$ 0.02} & 0.49 $\pm$ 0.02 & 0.42 $\pm$ 0.04 & 0.41 $\pm$ 0.03 & \textbf{0.74 $\pm$ 0.04} & 0.71 $\pm$ 0.03 & 0.62 $\pm$ 0.03 & 0.59 $\pm$ 0.04 \\
        & Macro & \textbf{0.50 $\pm$ 0.02} & 0.49 $\pm$ 0.02 & 0.43 $\pm$ 0.03 & 0.43 $\pm$ 0.03 & \textbf{0.70 $\pm$ 0.03} & 0.67 $\pm$ 0.04 & 0.58 $\pm$ 0.02 & 0.54 $\pm$ 0.04 \\ 
        \bottomrule
    \end{tabular}
    }
    \vspace{-1em}
    \label{tab:f1-scores-noisy-oracle}
\end{table}

Expectedly, introducing prediction noise into the target identities result in a modest decrease in absolute performance across all methods compared to the annotated-target setting. The relative performance across models still remains the same, with SAJ-MTL giving best results. On MHS, Joint-AWG achieves the highest macro-F1 of 0.59, outperforming both TextAug and CondMTL by a substantial margin while also obtaining the strongest micro-F1 performance. Similarly, on KET, the Joint-AWG variant consistently remains the best-performing model across all reported metrics, with Joint-AG also outperforming both baseline approaches. These results demonstrate that the benefits of explicitly modeling annotator-target interactions are preserved even when stakeholder identities are estimated imperfectly, indicating that SAJ-MTL has some degree of robustness to realistic levels of target detection noise.

\begin{table}[ht]
    \centering
    \vspace{-0.75em}
    \caption{Maximum EO Gap across demographic groups under noisy target prediction on MHS. Lower values indicate more equitable performance. Both Joint-AWG and Joint-AG maintain lower EO gaps than baseline methods despite using automatically predicted target identities.}
    \resizebox{0.7\textwidth}{!}{
    \begin{tabular}{l|cccc|c} \toprule
         \multirow{2}{*}{Model} & \multicolumn{4}{c|}{Recall} & Max. \\ 
         & Asian & Black & Latinx & White & EO Gap ($\downarrow$) \\ \midrule
         Joint-AWG & 0.49 $\pm$ 0.02 & 0.64 $\pm$ 0.02 & 0.44 $\pm$ 0.02 & 0.53 $\pm$ 0.02 & 0.20 $\pm$ 0.02 \\
        Joint-AG & 0.49 $\pm$ 0.02 & 0.64 $\pm$ 0.02 & 0.44 $\pm$ 0.02 & 0.50 $\pm$ 0.01 & 0.20 $\pm$ 0.02 \\
        TextAug & 0.38 $\pm$ 0.03 & 0.60 $\pm$ 0.02 & 0.35 $\pm$ 0.03 & 0.40 $\pm$ 0.02 & 0.25 $\pm$ 0.03 \\
        CondMTL & 0.30 $\pm$ 0.02 & 0.60 $\pm$ 0.02 & 0.30 $\pm$ 0.02 & 0.53 $\pm$ 0.02 & 0.30 $\pm$ 0.01 \\
        \bottomrule
    \end{tabular}
    }
    \vspace{-0.75em}
    \label{tab:eogap-noisy}
\end{table}

We further evaluate demographic fairness under the detected-target setting on MHS using the EO Gap measure, with results in \textbf{Table \ref{tab:eogap-noisy}}. Although imperfect target predictions do introduce additional uncertainty, both SAJ-MTL variants maintain lowest EO gaps. These results indicate that stakeholder-aware modeling remains effective even when demographic target identities are inferred automatically, and that fairness improvements under human-annotated target setting are not mere artifacts of perfect demographic information, but also translate to scenarios where target detection errors are unavoidable. 




\subsection{Scalability}

\textbf{Table \ref{tab:scalable-arch}} reports the memory (number of trainable parameters) and runtime (model training for 10 epochs) 
of our SAJ-MTL \vs other models. Target-Only CondMTL, which does not jointly model the annotator-target interactions, serves as baseline. TextAug requires the largest parameter and runtime among all the models. Being a text-augmentation model, the DistilBERT layers were made trainable to learn updated feature representation from the augmented post. These unfrozen DistilBERT layers are responsible for the massive increase in the parameter space. 
%
%
Our scalable SAJ-MTL model only has four annotator and four target branches corresponding to its additive architecture in Fig.~\ref{fig:sajmtl}. While maintaining comparable results \wrt naive SAJ-MTL model, this proposed framework scales well in terms of both memory and runtime. 
Even with the four demographic groups (\textit{Asian, Black, Latinx and White}) and two stakeholder (Annotator and Target) interactions, SAJ-MTL has $18\%$ fewer trainable parameters than naive SAJ-MTL. 
The $30\%$ runtime increase of SAJ-MTL 
is expected given its eight branches \vs only four in naive SAJ-MTL. From a computational viewpoint, as the group cardinality increases in any stakeholder, scalable SAJ-MTL can accommodate model training within the same machine, without GPU memory saturation. Additional simulation results of group scaling can be found in \textbf{Appendix~\ref{app:group-scaling}}. We note that AG and AWG represent different data protocols rather than different model architectures. Consequently, both variants have identical asymptotic computational and parameter complexity.


\begin{table}[ht]
\centering
\vspace{-0.5em}
    \caption{Memory (trainable parameters) and training time (10 epochs) comparison between models. We treat the Target-Only model as baseline for comparison, since it does not account for annotator views. While the model size of Target-Only and Naive SAJ-MTL are the same, SAJ-MTL takes longer time to train due to variation in data. The scalable SAJ variant is the best performing model in terms of trainable parameters.}
    \resizebox{0.65\textwidth}{!}{
	\begin{tabular}{l|rr|rr} \toprule
	    Model type & \# Params ($\downarrow$) & $\Delta$ ($\downarrow$) & Time(s) ($\downarrow$) & $\Delta$ ($\downarrow$) \\ \midrule
        Target Only (CondMTL) & 763,653 & - & 2200 & - \\ 
        TextAug & 67,075,795 & +8684\% & 4800 & +118\% \\ \midrule
        Naive SAJ-MTL & 763,653 & +0\% & 2400 & +9\% \\ 
        \midrule
	SAJ-MTL (Ours) & 626,133 & -18\% & 2850 & +30\% \\ 
    \bottomrule
	\end{tabular}
    }
    \vspace{-1em}
    \label{tab:scalable-arch}
\end{table}

\section{Discussion}

\textbf{Role of Disagreement in Annotation}. A recurring theme across subjective NLP tasks is that annotator disagreement is not merely noise but a reflection of societal and individual variation in lived experience. Literature increasingly emphasizes that collapsing multiple labels into a single ground truth can erase meaningful minority interpretations \citep{noble2012minority} and misrepresent the underlying construct being annotated. \citet{davani2022dealing} show that aggregating labels via majority vote systematically suppresses minority demographic variation, effectively erasing the valid perspectives of marginalized groups from the training signal. \citet{gordon2022jury} argue that heterogeneous judgments should not be discarded, as they encode essential information about how different demographic stakeholders distinctively perceive harm, offensiveness, and context. \citet{fleisig2023majority} highlight that relying on a forced consensus often leads models astray; they show that retaining un-aggregated annotations allows classifiers to learn robustly even when the majority opinion is contextually flawed or socially biased. The proposed methodology contributes to this body of work through a scalable multi-task learning framework for modeling disagreement in annotation tasks, which demonstrates significant improvements for modeling toxicity detection.


\textbf{Interpreting Viewpoint Prioritization in Practice.} While our model can represent a wide array of demographic perspectives, the decision of which viewpoint to prioritize remains a task-specific normative choice. In SAJ-MTL, this prioritization is formalized through a weighting function $\pi(a,t)$, reflecting epistemic, pluralistic, policy-driven, or task-specific decisions about whose perspective should matter most. Following the perspective-aware framework of \citet{basile2021we}, our work supports a shift beyond a single gold standard toward a system that accounts for disaggregated human perspectives. The flexible design of $\pi(a,t)$ allows practitioners to encode diverse priorities at inference time, integrating annotators' group membership as a primary component of the model’s prediction logic rather than treating it solely as a feature of data collection. This approach directly addresses the findings of \citet{prabhakaran2021releasing}, who demonstrated that standard label aggregation (like majority voting) often introduces systematic biases by over-representing majority perspectives and obscuring the lived experiences of minority groups. In our framework, prioritization is not used to correct the model, but to enable explicit normative choices  of how different voices are accounted for. For example, it can be used to center the voices of those with the highest stake in the outcome. In the context of anti-Asian hate speech, for instance, the model can be configured to prioritize the perspective of Asian annotators, ensuring the detection threshold aligns with the community most directly impacted. This flexibility enables a more equitable and context-aware moderation strategy that adapts to different normative choices without requiring retraining. SAJ-MTL thus operationalizes the principle that interpretation depends not only on what is being annotated, but on who is annotating and whom the content targets, without requiring policy priorities to affect the training pipeline.


\textbf{Generalization to other tasks.} Although in this work we explicitly focused on toxicity detection, the SAJ-MTL framework naturally generalizes to many subjective or context-sensitive tasks in NLP and beyond. Tasks such as stance classification, sentiment analysis, misinformation verification, and even medical image analysis have well-documented sources of demographic variation in annotation. For instance, 
in medical image analysis, diagnoses can vary systematically based on a clinician's training, experience, and the demographic presentation of the patient. Rather than treating such clinical disagreement as noise, SAJ-MTL could be used to factorize these expertise and patient-context variables to preserve critical diagnostic signals \citep{zhong2026learning}. As noted by \citet{sorensen2024roadmap}, as AI systems are increasingly deployed in global contexts, they must move away from average human preferences toward a value-pluralistic stance. Because SAJ-MTL’s conditional structure simply combines shared semantic encoders with annotator- and target-specific branches, it does not assume anything special about toxicity labels and can be applied wherever group-conditioned interpretation is relevant. Moreover, because inference-time aggregation is decoupled from model training, domain experts in these diverse fields can tailor viewpoint prioritization to reflect domain-specific norms, regulatory structures, or governance criteria without retraining the underlying model.






\section{Conclusion}


Human annotations in subjective tasks often reflect socially grounded disagreement rather than objective ground truth. Single-truth or majority-voted paradigms treat such variation as noise, discarding valuable signal about annotator and target-group perspectives. To address this, we introduced SAJ-MTL, a joint annotator–target–aware multi-task learning framework that explicitly models interactions between annotators' demographics and content. Using a factorized conditional architecture with selective gradient routing, SAJ-MTL captures structured cross- and within-group variation while scaling linearly with the number of groups and requiring only minimal group-membership information in order to preserve privacy. A flexible viewpoint prioritization mechanism further decouples disagreement modeling from deployment-time choices, allowing practitioners to encode epistemic authority, pluralism, or policy constraints without retraining. Applied to toxicity detection, SAJ-MTL improves macro-level performance on demographic-sensitive content and reduces group-level disparities, all without explicit fairness objectives.

Beyond toxicity, SAJ-MTL offers a general template for learning from subjective annotations where stakeholder perspectives vary systematically. The additive factorization of annotator and target groups suggests a natural extension to a stakeholder triad that incorporates author demographics, as recent works show that author dialect and identity also shape toxicity perception \citep{chaudhry2022you,miao2023detecting}. Integrating such signals additively would capture triadic dynamics while maintaining our privacy-preserving, group-level formulation. More broadly, future work could relax rigid group categories by learning hierarchical or latent group structures, enabling the model to discover intersectional identities obscured by broad demographic labels. Finally, SAJ-MTL's architecture is well-suited to multi-modal applications: applied to medical imaging with a Vision Transformer encoder, it could reconcile systematic diagnostic disagreements across clinical traditions or specialties, by treating divergent expert judgments as signal rather than noise.

\section{Ethical Considerations}

The ethical stakes of modeling annotator disagreement are significant. Without explicit design choices, ML systems risk implicitly weighing majority viewpoints, thereby amplifying harms to marginalized or under-represented groups and reinforcing pre-existing biases~\citep{sap2019risk,davidson2019racial}. 
Still, decisions around how groups are defined, aggregated, or weighted require careful ethical consideration. Across application domains, anchoring these decisions on domain-knowledge, and engaging in participatory design practices may be important avenues to ensure their relevance and legitimacy. Additionally, it is important to note that there is a risk of misuse. It could be possible to define viewpoint prioritization functions that perpetuate or exacerbate marginalization by purposefully centering privileged viewpoints. Thus, transparency regarding viewpoint prioritization mechanisms deployed is essential.

The proposed methodology does not explicitly enforce fairness. Instead, fairness benefits stem from improvements in modeling. This is, however, not guaranteed to happen, and it is essential to explicitly evaluate fairness performance, as illustrated in our experiments. In sensitive domains such as toxicity detection, minimizing performance gaps is not merely an algorithmic optimization metric, but a necessary safeguard against asymmetric harms. Errors may disproportionately affect i) the author of content, via unfair over-moderation or mischarecterization, and ii) the communities targeted by that content, via failure to filter out or appropriately react to harmful content. Careful evaluation across stakeholder groups is therefore required and necessary to avoid reinforcing existing social inequities. 

Our experiments rely exclusively on publicly available datasets \cite{sachdeva2022assessing,kumar2021designing}, and no additional human participants were recruited or annotations collected. We use the datasets subject to their appropriate access and usage conditions. The use of annotator demographic information introduces privacy and representation concerns. Although our methodology operates on group-level demographic attributes, such information may still be sensitive in some scenarios \cite{sweeney2013discrimination,barocas2016big,veale2017fairer}, and its use in deployed ML systems should therefore be subject to appropriate privacy guardrails, informed consent where applicable, and restrictions against discriminatory profiling.

\section*{Acknowledgment}

We thank the data annotators who made this work possible, as well as our anonymous reviewers for their valuable feedback. This research was supported in part by Amazon, Wipro, the Knight Foundation, the Micron Foundation, and by Good Systems\footnote{\url{https://goodsystems.utexas.edu/}}, a UT Austin Grand Challenge to develop responsible AI technologies. The statements made herein are solely the opinions of the authors and do not reflect the views of the sponsoring agencies.



\clearpage

\bibliographystyle{ACM-Reference-Format}
\bibliography{References}

\appendix

\section{SAJ-MTL as a Neural Generalization of Identity-Sensitive IRT} \label{app:irt-linkage}

This section provides a psychometric interpretation of SAJ-MTL through the lens of identity-sensitive Item Response Theory (IRT) \citep{sachdeva2022assessing}. We show that the proposed architecture preserves the additive inductive structure underlying latent trait models while extending them into a scalable neural conditional representation framework over annotator-target demographic spaces. We also note that while classical IRT formulations such as the 2PL model express probability multiplicatively as $P = \sigma(a(\theta - \beta))$, the identity-sensitive IRT variant of \citet{sachdeva2022assessing} operates in the log-odds (latent response) space, where item, annotator, and demographic factors enter additively. SAJ-MTL preserves this additive inductive bias, extending it from scalar latent variables to high-dimensional neural representations.

The IRT framework models annotation behavior through additive latent response factors corresponding to item characteristics and annotator-dependent perception effects. Their final formulation defines:
\begin{equation}
    \log\left[\frac{p_{nijk}}{p_{nij(k-1)}}\right]=\theta_n - \delta_i - \alpha_j - \tau_k + \beta_{jm} + \mu_{lm}
\end{equation}
where $\theta_n$ denotes the latent toxicity tendency of item $n$, $\delta_i$ represents item difficulty, $\alpha_j$ captures annotator-specific sensitivity or severity, $\tau_k$ denotes ordinal threshold parameters, $\beta_{jm}$ captures whether annotator $j$ exhibits increased or decreased sensitivity toward content targeting demographic group $m$, and $\mu_{lm}$ captures systematic alignment effects between annotator identity $l$ and targeted demographic identity $m$.

SAJ-MTL follows an analogous additive inductive bias, but replaces scalar latent variables with learned high-dimensional neural representations. Specifically, we define a task-conditional prediction function $f_\theta(y \mid x,a,t)$ jointly on: (i) input text $x$, (ii) annotator demographic group $a$, and (iii) targeted demographic group $t$.

The shared conditional representation $z(x,a,t)$ is instantiated through the additive parameterization:
\begin{equation*}
    z(x,a,t)=h_\phi(x) + W_a e_a + W_t e_t
\end{equation*}
Under this interpretation, the shared encoder representation $h_\phi(x)$ generalizes latent item-dependent effects $(\theta_n,\delta_i)$, the annotator component $W_a e_a$ generalizes annotator sensitivity effects $(\alpha_j)$, and the target component $W_t e_t$ introduces demographic-target conditioning absent from classical IRT formulations.

Thus, SAJ-MTL preserves the additive structure of latent trait models while extending them from scalar psychometric variables to structured high-dimensional neural conditional representations.

The interaction-sensitive term $\beta_{jm}$ aligns closely with the demographic-conditioned prediction structure of SAJ-MTL
$f_\theta(y \mid x,a,t)$, and in particular with the additive parameterization, $ W_{a,t} = U_a + V_t$, where $U_a$ models systematic annotator-group response tendencies, and $V_t$ models target-group-specific toxicity sensitivity. While $\beta_{jm}$ represents scalar interaction adjustments in a probabilistic latent trait framework, SAJ-MTL generalizes this interaction sensitivity into high-dimensional conditional neural representations.

The term $\mu_{lm}$ captures alignment effects between annotator identity and targeted demographic identity. Of particular importance are self-interactions $(l=m)$, corresponding to annotators evaluating content targeted toward their own demographics ($a=t$). This directly parallels the epistemic authority weighting formulation used in SAJ-MTL as $\pi(a,t)$, which models evaluation according to annotator-target demographic alignment.

\begin{table}[ht]
    \centering
    \vspace{-0.5em}
    \caption{Stage-wise correspondence between identity-sensitive IRT formulations and SAJ-MTL.}
    \resizebox{\textwidth}{!}{
    \begin{tabular}{lll}
    \toprule
    \textbf{IRT Component} & \textbf{Interpretation} & \textbf{SAJ-MTL Correspondence} \\
    \midrule
    $\theta_n,\delta_i$ & Item toxicity and difficulty & Shared encoder representation $h_\phi(x)$ \\
    $\alpha_j$ & Annotator-specific sensitivity & Annotator component $U_a$ \\
    $\beta_{jm}$ & Annotator--target interaction sensitivity & Conditional demographic predictor $f_\theta(y \mid x,a,t)$ \\
    $\mu_{lm}$ & Group-level demographic alignment & Epistemic authority weighting $\pi(a,t)$ \\
    Eq.~\ref{eq:fcond-add} & Additive latent response decomposition & Shared conditional representation learning \\
    Eq.~\ref{eq:ann-tar-logits} & Interaction-sensitive demographic modeling & Demographic-conditioned prediction parameterization \\
    Eq.~\ref{eq:pi} & Group-level viewpoint weighting & Viewpoint-prioritized inference aggregation \\
    \bottomrule
    \end{tabular}
    }
    \vspace{-0.5em}
    \label{tab:irt-saj}
\end{table}

A key distinction, however, is that identity-sensitive IRT incorporates these effects directly within probabilistic parameter estimation, whereas SAJ-MTL decouples demographic weighting from representation learning. Consequently, alternative viewpoint prioritization schemes may be applied post hoc during evaluation without retraining the underlying predictive model. Taken together, SAJ-MTL may be interpreted as a structured neural extension of identity-sensitive IRT modeling, where scalar latent demographic effects are generalized into high-dimensional conditional representations and modular viewpoint-aware inference mechanisms. A mapping correspondence of different stages of SAJ-MTL to the IRT formulation is given in \textbf{Table \ref{tab:irt-saj}}.

\section{Experimental Details} \label{app:details}

\textbf{Setup.} Experiments use a Nvidia 2060 RTX Super 8GB GPU, Intel Core i7-9700F 3.0GHz 8-core CPU and 16GB DDR4 memory. We use Keras \citep{chollet2015} library on a Tensorflow 2.0 backend with Python 3.12 to train the networks in this paper. For optimization, we use AdaMax \citep{kingma2014adam} with parameters (\textit{lr}=0.001) and $1000$ steps per epoch. We also setup an EarlyStopping criteria in case the with a min\_delta of $1\textsc{e-}4$ and patience of 5. The random seed was set to 42 to avoid varaibility.

\textbf{Optimizer.} We choose AdaMax (infinity-norm variant of Adam) due to its faster convergence from fresh starts, \ie model training from random weights. It is also more stable than Adam in our case with sparse or unbalanced gradients. This makes tuning the learning rate less fragile.

\textbf{Regularization.} We do not need any explicit regularization term in our SAJ-MTL loss formulation, since all the task branches are treated equally and each branch has a conditional binary cross entropy loss acting on it. For branches having extreme sparsity, we have independent $\ell_2$ regularization for smoothness, which we have explicitly set to $\lambda_{a,t}=0.005$. However, during our epistemically weighted evaluation over both datasets, this additional term and its variation did not cause any metrics to fall outside the reported variance range.

\textbf{Training protocol}: To ensure a fair comparison with SAJ-MTL, all models are trained using the same shared encoder architecture and optimization procedure. 
All models are trained on individual annotations rather than majority-vote labels, ensuring that disagreement signals are preserved across methods. For SAJ-MTL, annotator and target embeddings are initialized randomly and projected into the shared representation space using linear transformations $W_a,W_t$ before being combined with the shared representation. Predictions are then produced through the additive combination of annotator-specific and target-specific heads as described in Section \ref{sec:arch}. By comparing these approaches, we evaluate whether explicitly modeling the interaction between annotator and target demographics, rather than incorporating metadata implicitly or modeling only target groups, can achieve improved predictive performance and fairness.

\section{Dataset Details} \label{app:dataset}

We use the Measuring Hate Speech (MHS) corpus \citep{sachdeva2022assessing}, available via the HuggingFace library \citep{huggingfaceDlab}, which includes 135,556 posts sourced from YouTube, Twitter and Reddit. Each post is annotated by one or more human annotators and includes: i) the inferred demograhic group(s) targeted by the post; ii) self-reported group-demographic attributes of the annotators; and iii) a binary toxicity label indicating whether the content is perceived as \textit{toxic} or \textit{non-toxic} towards the inferred target. Although the dataset has seven demographic race groups as target, we keep only four demographic groups - \textit{Asian}, \textit{Black}, \textit{Latinx} and \textit{White}, and omit three others (\textit{Middle-Eastern}, \textit{Native-American} and \textit{Pacific-Islander}). This is due to extremely sparse annotator–target ($A_i,T_j$) pair instances in the three omitted groups. For included groups, both annotators and targets were drawn as shown in \textbf{Table \ref{tab:hugg-joint-stats}}. Because individual posts may be annotated by multiple annotators, we perform the train–validation–test split at the post level to prevent content leakage across splits. Specifically, posts are randomly partitioned into 70\%, 10\%, and 20\% subsets for training, validation, and testing, respectively. All annotations associated with a given post are assigned to the same split. Within each split, a post may still have multiple annotator labels, which allows us to model annotator-specific variation while preserving strict separation of textual content across splits. The dataset exhibits substantial variation in the frequency with which different demographic groups are targeted. For example, the Black community represents the statistical majority among targeted groups.

\begin{table}[h]
    \centering
    \caption{MHS corpus statistics showing the number of posts that match a specific annotator-target pair ($A_i,T_j$) over four demographic groups. Cells highlighted in green denote the ones considered under epistemic authority weighting.}
    \resizebox{0.75\textwidth}{!}{
    \begin{tabular}{r|rrrr|r}
    \toprule
    Target & \multicolumn{4}{c|}{Annotator Group} & All Annotators \\
    Group & Asian & Black & Latinx & White & \\ \midrule
    Asian & \cellcolor{green!50}621 (1.2\%) & 721 (1.4\%) & 551 (1.1\%) & 5,501 (10.8\%) & 7,394 (14.6\%) \\
    Black & 1,560 (3.1\%) & \cellcolor{green!50}2,292 (4.5\%) & 1,654 (3.3\%) & 18,545 (36.6\%) & 24,051 (47.4\%) \\
    Latinx & 638 (1.3\%) & 961 (1.9\%) & \cellcolor{green!50}700 (1.4\%) & 6,657 (13.1\%) & 8,956 (17.6\%) \\
    White & 711 (1.4\%) & 1,086 (2.1\%) & 706 (1.4\%) & \cellcolor{green!50}7,825 (15.4\%) & 10,328 (20.4\%) \\ \midrule
    All Targets & 3,530 (6.7\%) & 5,060 (10.0\%) & 3,611 (7.1\%) & 38,528 (75.9\%) & 50,729 (100\%) \\
    \bottomrule
    \end{tabular}
    }
    \label{tab:hugg-joint-stats}
    \vspace{-0.5em}
\end{table}

The KET dataset \citep{kumar2021designing} consists of posts sourced from Twitter, Reddit, and 4chan. We focus on four annotator groups based on race demographics: \textit{Asian}, \textit{Black}, \textit{Latino}, and  \textit{White}. Each annotator has marked the post as being non-toxic (0) or toxic (1-4), which we binarize as $y \in \{0,1\}$ for our formulation. Because target groups are not annotated for this dataset, one must detect targets automatically, as in~\citet{fleisig2023majority}. Again, due to lack of pre-defined splits, we randomly partitioned posts into 70\%, 10\%, and 20\% subsets for training, validation, and testing, respectively. In this dataset, the White group has the majority of posts, as shown in \textbf{Table \ref{tab:ket-joint-stats}}.

\begin{table}[h]
    \centering
    \caption{KET statistics showing posts that match a specific annotator ($A_i$) over four demographic groups.}
    \resizebox{0.75\textwidth}{!}{
    \begin{tabular}{r|rrrr|r}
    \toprule
    Target & \multicolumn{4}{c|}{Annotator Group} & All Annotators \\
    Group & Asian & Black & Latinx & White & \\ \midrule
    Unmarked & 33,140 (15.3\%) & 56,580 (26.1\%) & 19,500 (9.0\%) & 107,480 (49.6\%) & 216,700 (100\%) \\
    \bottomrule
    \end{tabular}
    }
    \label{tab:ket-joint-stats}
\end{table}

\section{Automatic Target Detection} \label{app:noisy-oracle}

In many real-world deployment settings, the demographic group targeted by a post may not be explicitly available and must instead be inferred automatically. While MHS provides human-annotated target-groups, such targets will typically need to be automatically (noisily) detected. Consequently, stakeholder-aware prediction systems often rely on an auxiliary model to estimate target identities before downstream classification. We perform automatic target detection to evaluate SAJ-MTL under this more realistic setting, to assess the robustness of stakeholder-aware modeling when demographic target assignments contain errors. 

Specifically, let $t_{ann}$ denote the annotated gold target demographic label associated with a sample. Instead of providing $t_{ann}$ directly to the toxicity classifier, we replace it with a predicted target label $t_{pred}$ obtained from an auxiliary target-group classifier as $t_{pred} = f_{target}(x)$, where $f_{target}$ predicts one or more of the demographic target categories from the text alone. During both training and evaluation of downstream toxicity models, all methods receive $t_{pred}$ rather than the annotated target label $t_{ann}$.

For MHS, we ignore gold labels and use predicted labels. 
For KET, target identities are not annotated. 

\begin{figure}[ht]
    \centering
    \vspace{-1em}
    \includegraphics[width=0.6\linewidth]{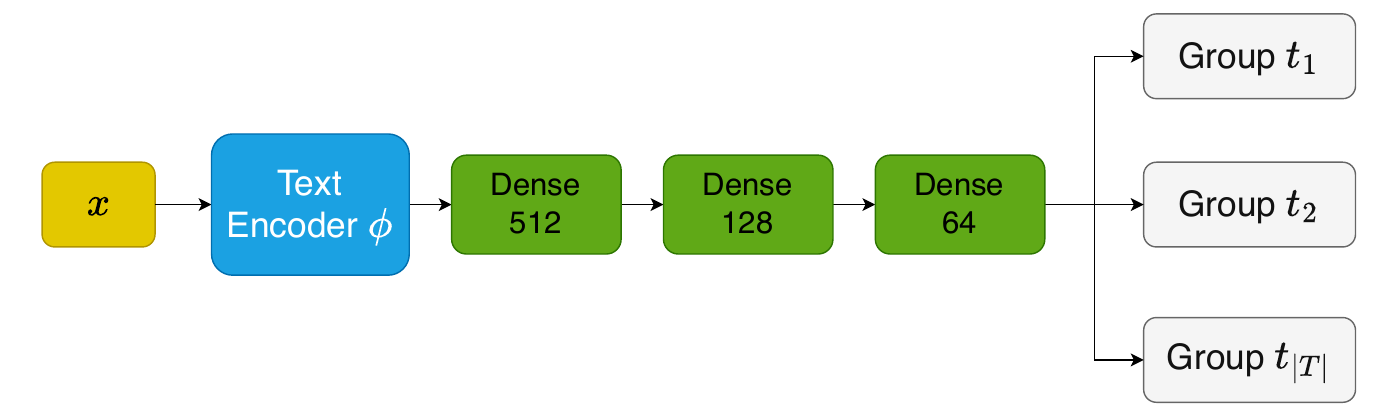}
    \vspace{-0.75em}
    \caption{Target-group detection model. The final layer has $|T|$ nodes, each containing a classification layer.}
    \vspace{-1em}
    \label{fig:target-module}
\end{figure}

Similar to the protocol by \citet{fleisig2023majority}, we use an custom target detection model. For their work, they used a GPT-2 based module fine-tuned on a different dataset. However, note that different platforms and sources of data come with their own length formats, audience demographics, moderation policies \etc. Hence to prevent such artifacts originating from decoupling, our custom target detection model uses the same underlying text encoder architecture as the downstream models and is trained independently using available target annotations where applicable \ie the MHS dataset in our case. This makes the entire pipeline coupled in terms of data source and text representation. To avoid information leakage, target detection and downstream toxicity classification are trained separately on the same splits. The text encoder is followed by dense neuron layers (512, 128, 64 neurons respectively), with \textit{relu} activation, added biases and dropout rate of 0.1 to prevent model overfitting. The layers end in $|T|$ classification nodes, each with \textit{sigmoid} activation, no biases and 0.5 threshold. Given that a single post can target one or more groups simultaneously, we use weighted Binary Cross Entropy (wBCE) as our loss to optimize overall error. The rounded detection accuracy rates for the demographic groups were as follows: Asian: 81\%, Black: 87\%, Latinx: 82\%, White: 77\%, with the corresponding confusion matrix shown in \textbf{Table \ref{tab:conf-mat-mhs}}.

\begin{table}[ht]
    \centering
    \caption{Confusion matrix (counts) for MHS between using the target-group detection model.}
    \resizebox{0.5\textwidth}{!}{
    \begin{tabular}{cc|rrrr} \toprule
         & & \multicolumn{4}{c}{Detected Target Group} \\ \cmidrule{3-6}
        & & Asian & Black & Latinx & White \\ \midrule
        \multirow{4}{*}{\parbox{1.6cm}{\centering Annotated\\ Target\\ Group}} & Asian & 5,989 & 151 & 107 & 1,147 \\
        & Black & 242 & 20,924 & 247 & 2,638 \\
        & Latinx & 121 & 173 & 7,344 & 1,318 \\
        & White & 687 & 985 & 703 & 7,953 \\
        \bottomrule
    \end{tabular}
    }
    \vspace{-1em}
    \label{tab:conf-mat-mhs}
\end{table}

\section{Additional Results} \label{app:roberta}

\begin{table}[h]
    \centering
    \vspace{-0.5em}
    \caption{Performance on MHS in the Annotated-Target Scenario using RoBERTa-base as text encoder. Mean model performance and deviations are reported over 5 runs. Best performance is highlighted in \textbf{bold}.}
    \resizebox{0.65\textwidth}{!}{
    \begin{tabular}{r|c|cc|cc} 
    \toprule 
    Metric & Level & Joint-AWG & Joint-AG & TextAug & CondMTL \\ \midrule 
    F1 ($\uparrow$) & Micro & \textbf{0.67 $\pm$ 0.01} & 0.63 $\pm$ 0.02 & 0.56 $\pm$ 0.02 & 0.52 $\pm$ 0.02 \\ 
    & Macro & \textbf{0.70 $\pm$ 0.02} & 0.66 $\pm$ 0.01 & 0.57 $\pm$ 0.02 & 0.50 $\pm$ 0.01 \\ \midrule 
    Precision ($\uparrow$) & Micro & \textbf{0.75 $\pm$ 0.02} & 0.70 $\pm$ 0.01 & 0.66 $\pm$ 0.03 & 0.62 $\pm$ 0.02 \\ 
    & Macro & \textbf{0.78 $\pm$ 0.01} & 0.74 $\pm$ 0.02 & 0.66 $\pm$ 0.03 & 0.60 $\pm$ 0.02 \\ \midrule 
    Recall ($\uparrow$) & Micro & \textbf{0.61 $\pm$ 0.01} & 0.58 $\pm$ 0.02 & 0.48 $\pm$ 0.02 & 0.45 $\pm$ 0.01 \\ 
    & Macro & \textbf{0.63 $\pm$ 0.02} & 0.60 $\pm$ 0.02 & 0.50 $\pm$ 0.02 & 0.43 $\pm$ 0.02 \\ 
    \bottomrule 
    \end{tabular}
    }
    \label{tab:joint-summary-mhs-roberta}
    \vspace{-0.5em}
\end{table}

\textbf{Table \ref{tab:joint-summary-mhs-roberta}} presents the statistics on MHS while switching from DistilBERT to RoBERTa-base as the input text encoder with frozen weights. It is expected that RoBERTa's enhanced capabilities could further improve these scores due to its larger size and advanced pre-training strategies. The results indicate that Joint-AWG and Joint-AG consistently outperforms the other baseline methods in both Micro and Macro evaluations for all metrics. This suggests that SAJ-MTL variants effectively balances precision and recall, demonstrating high accuracy and relevance in its predictions by accounting for demographic disagreements. The Joint-AWG is still the best model by considering within and across group variations. This analysis not only highlights the modeling strength, but also potential improvements with powerful encoders, weighing against computational trade-offs for practical deployment considerations.

\section{Scalability of SAJ-MTL} \label{app:group-scaling}

A key design goal of SAJ-MTL is to scale nominally with increasing numbers of annotator and target groups. However, in the naive design as shown in Fig.~\ref{fig:naive}, this leads to multiplicative parameter growth of $\mathcal{O}(|A| \times |T|)$. SAJ-MTL addresses this challenge through a factorized parameterization in which annotator and target effects are learned independently and composed additively. With a single shared encoder, a small set of annotator and target embeddings, the total number of group-specific parameters scales additively as ($\mathcal{O}(|A| + |T|)$) (Fig.~\ref{fig:scalable}). This ensures tractability even in settings with dozens of annotator groups or highly diverse target identities. Concretely, rather than instantiating a separate head or parameter block for each $(a,t)$ pair, SAJ-MTL reuses a single shared encoder and a set of small annotator and target modules, allowing the model to represent structured disagreement without incurring combinatorial cost.

In practice, this design leverages parameter-efficient model design ideas from \citep{houlsby2019parameter,hu2022lora}. Beyond memory scalability, this parameterization enhances computational efficiency. Because only the annotator and target modules relevant to an example receive gradient updates, SAJ-MTL amortizes shared backbone updates and computation across tasks. By conditioning predictions on small learned embeddings rather than duplicating large network components, batch-level throughput remains stable even as the number of groups grows. Compared to frameworks such as Mixture-of-Experts \citep{shazeer2017outrageously,ma2018modeling}, which introduce routing and communication overhead, SAJ-MTL achieves scalability with minimal architectural overhead for embedding lookup and head routing. This design ensures fairness-aware modeling remains deployable in resource-constrained environments. Finally, additive factorization promotes generalization across sparse groups. Groups with limited labeled examples benefit from the shared encoder while still receiving individualized modeling capacity. Conversely, well-represented groups can develop more specialized representations through repeated branch-specific updates. This makes SAJ-MTL robust to the long-tailed, heterogeneous distributions common in subjective labeling settings.

\begin{figure}[htb!]
    \centering
    \vspace{-0.5em}
    \includegraphics[width=\linewidth]{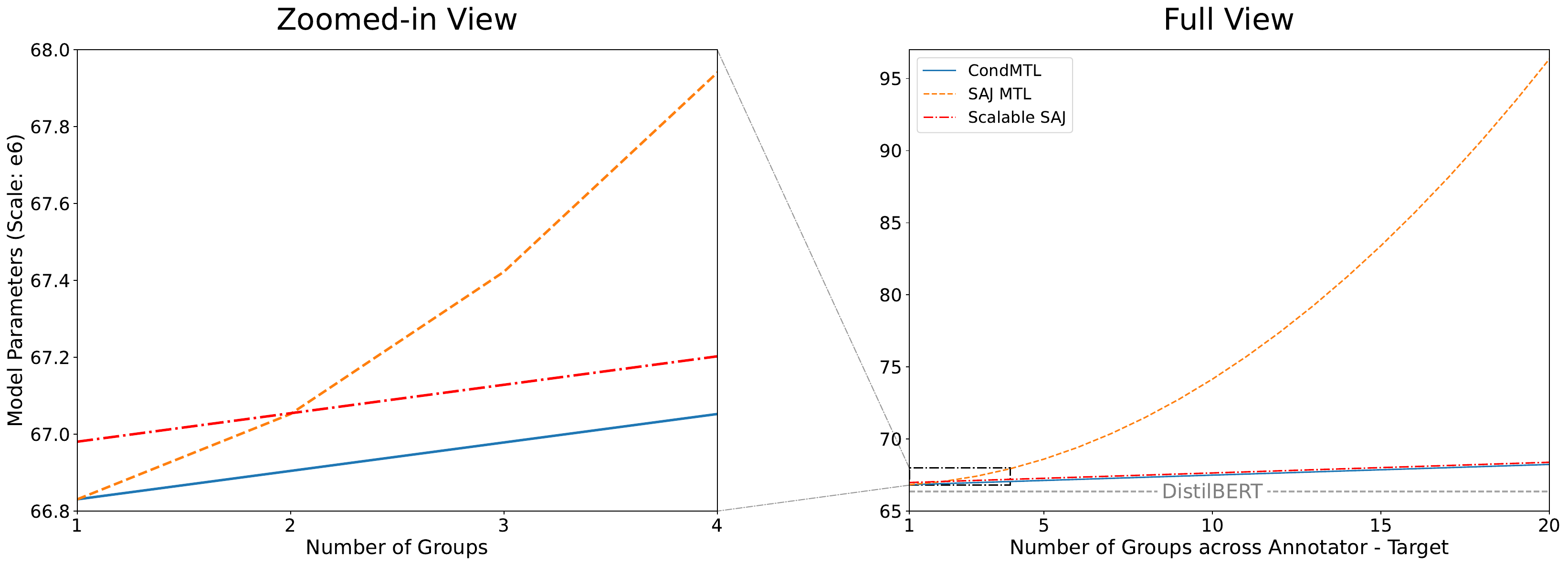}
    \vspace{-1em}
    \caption{Synthetic experiment on scalability. Every model has a fixed frozen (non-trainable) DistilBERT representation as their minimum memory. SAJ MTL's quadratic scaling nature with increasing groups is evident, while the Scalable variant is much more memory efficient due to having linear scaling with groups.}
    \vspace{-0.75em}
    \label{fig:scaling}
\end{figure}

\textbf{Fig.~\ref{fig:scaling}} shows a simulation of memory scaling across three models \vs some annotator-target pairings. All models have a fixed DistilBERT weight as minimum memory and additional dense layers. When the number of groups is one, we see that CondMTL and SAJ-MTL have effectively the same memory, since they end up with the same architecture. With increasing groups, the quadratic scaling kicks in and we see the memory for naive SAJ-MTL quickly saturating towards GPU limit, a phenomena exhibited by any standard MTL architecture. However, the Scalable SAJ-MTL variant almost grows asymptotically \wrt CondMTL (which just considers target group), all the while capturing interactions across both annotator-target groups.

\end{document}